\documentclass[letterpaper]{article} 
\usepackage{aaai2026}  
\usepackage{times}  
\usepackage{helvet}  
\usepackage{courier}  
\usepackage[hyphens]{url}  
\usepackage{graphicx} 
\usepackage{natbib}  
\usepackage{caption} 
\usepackage{algorithm}
\usepackage{algorithmic}

\usepackage{newfloat}
\usepackage{listings}

\usepackage{algorithm}
\usepackage{algorithmic}
\usepackage{graphicx}
\usepackage{amsmath}
\usepackage{amsthm}
\usepackage{multirow}
\usepackage{longtable}
\usepackage{bm}
\usepackage{amssymb}
\usepackage{subcaption}
\usepackage{xcolor}
\usepackage{colortbl}

\DeclareCaptionStyle{ruled}{labelfont=normalfont,labelsep=colon,strut=off} 
\floatstyle{ruled}
\newfloat{listing}{tb}{lst}{}
\floatname{listing}{Listing}
\title{
NumCoKE: Ordinal-Aware Numerical Reasoning over Knowledge Graphs with Mixture-of-Experts and Contrastive Learning
}
\author {
    Ming Yin\textsuperscript{\rm 1,2,3}, 
    Zongsheng Cao\textsuperscript{\rm 4}, 
    Qiqing Xia\textsuperscript{\rm 1,2,3}, 
    Chenyang Tu\textsuperscript{\rm 1,2}, 
    Neng Gao\textsuperscript{\rm 1,2,\textdagger}
}
\affiliations {
    \textsuperscript{\scriptsize 1}Institute of Information Engineering,Chinese Academy of Sciences.\\
    \textsuperscript{\scriptsize 2}State Key Laboratory of Cyberspace Security Defense.\\
    \textsuperscript{\scriptsize 3}School of Cyber Security, University of Chinese Academy of Sciences.\\
    \textsuperscript{\scriptsize 4}Department of Information Science, Tsinghua University.\\
    agiczsr@gmail.com, \{yinming, xiaqiqing, tuchenyang, gaoneng\}@iie.ac.cn\\
}

\usepackage{bibentry}

\begin{document}

\maketitle

\begin{abstract}
Knowledge graphs (KGs) serve as a vital backbone for a wide range of AI applications, including natural language understanding and recommendation. A promising yet underexplored direction is \textit{numerical reasoning} over KGs, which involves inferring new facts by leveraging not only symbolic triples but also numerical attribute values (e.g., \textit{length}, \textit{weight}). 
However, existing methods fall short in two key aspects:  
(1) \textbf{Incomplete semantic integration}: Most models struggle to jointly encode entities, relations, and numerical attributes in a unified representation space, limiting their ability to extract relation-aware semantics from numeric information.  
(2) \textbf{Ordinal indistinguishability}: Due to subtle differences between close values and sampling imbalance, models often fail to capture fine-grained ordinal relationships (e.g., longer, heavier), especially in the presence of hard negatives.
To address these challenges, we propose \textbf{NumCoKE}—a numerical reasoning framework for KGs based on Mixture-of-Experts and Ordinal Contrastive Embedding. To overcome (C1), we introduce a Mixture-of-Experts Knowledge-Aware (MoEKA) encoder that jointly aligns symbolic and numeric components into a shared semantic space, while dynamically routing attribute features to relation-specific experts. To handle (C2), we propose Ordinal Knowledge Contrastive Learning (OKCL), which constructs ordinal-aware positive and negative samples using prior knowledge, enabling the model to better discriminate subtle semantic shifts.
Extensive experiments on three public KG benchmarks demonstrate that \textbf{NumCoKE} consistently outperforms competitive baselines across diverse attribute distributions, validating its superiority in both semantic integration and ordinal reasoning.
\end{abstract}


\section{Introduction}
Knowledge graphs (KGs) represent structured factual knowledge as triples of entities and relations, and have become a key foundation for various AI applications, including recommendation systems~\cite{DBLP:conf/mm/CaoSW0WY22,DBLP:conf/mm/LiXJCH20}, natural language processing~\cite{DBLP:conf/mm/Guo0ZM0WX23,DBLP:conf/mm/LuDYY23,DBLP:conf/mm/SunYLLLS23}, and multimodal tasks~\cite{DBLP:conf/mm/WangMCML023,DBLP:conf/mm/LiQZ0TXT23}. A crucial yet underexplored capability of KGs is \textit{numerical reasoning}, the ability to infer facts involving quantitative comparisons or numerical ordering (e.g., \textit{“the Nile is longer than the Amazon”}). This ability is particularly valuable for information services that require precise numerical judgments, such as fine-grained product recommendations or numerical question answering.

\begin{figure}[t]
    \centering
    \includegraphics[scale=0.468]{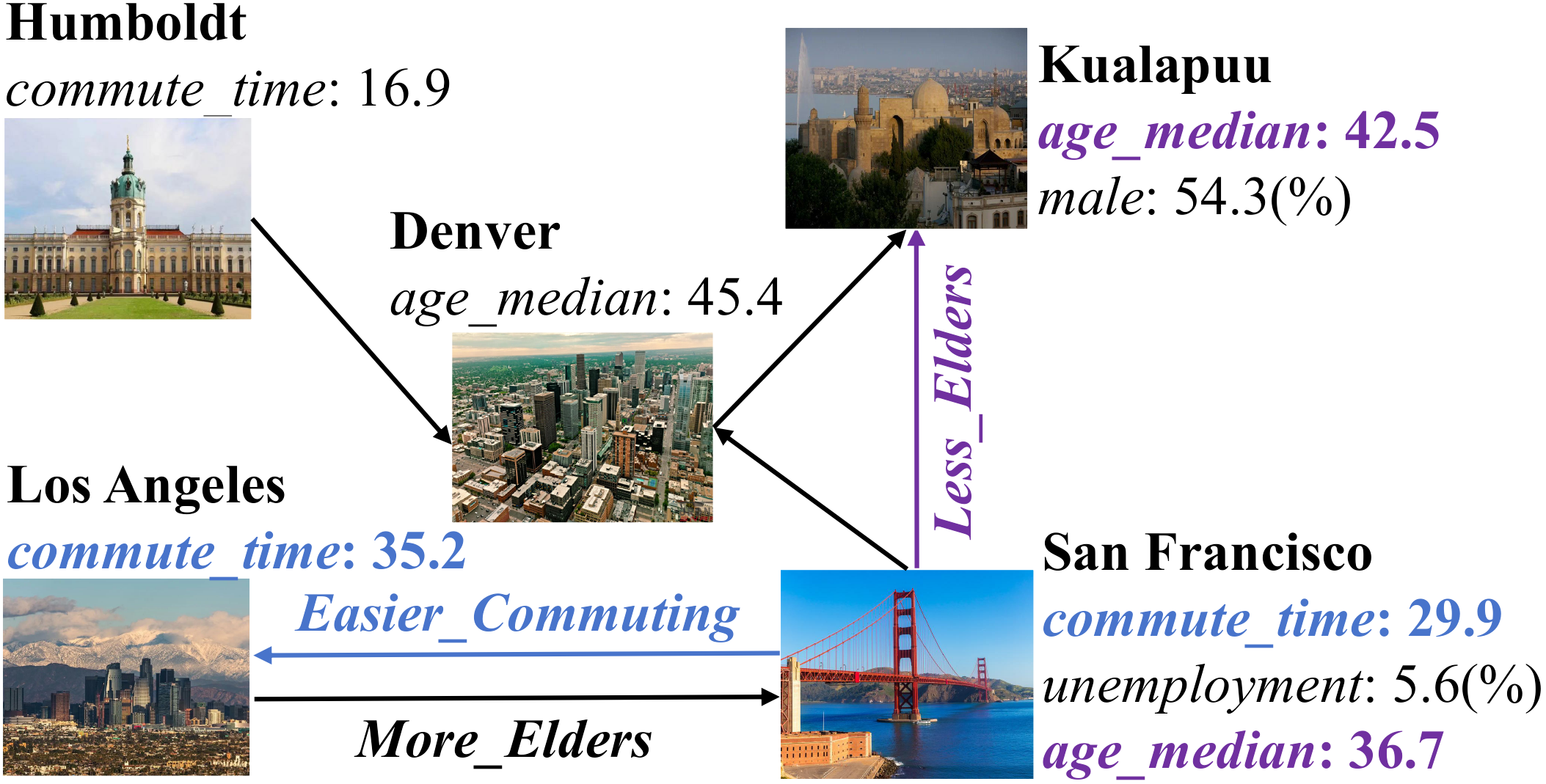}
    \caption{Example of KG-based numerical reasoning. Same colored text indicates strong related relation and attribute.}
    \label{NBA-KG}
\end{figure}

To enhance numerical reasoning, prior studies have explored integrating attribute values into knowledge graph embedding (KGE) frameworks. Some directly embed numerical entities into continuous spaces~\cite{DBLP:conf/kdd/BaiLLYYS23}, while others employ graph neural networks~\cite{DBLP:conf/iclr/VashishthSNT20} or augment traditional KGE with attribute-aware components~\cite{DBLP:conf/kdd/KimKKPJP23}. Despite these advances, existing methods face two persistent and intertwined challenges:
\begin{enumerate}
    \item[\textbf{C1}] \textbf{Incomplete semantic integration.} Current models often fail to effectively capture the joint semantics of entities, relations, and numerical attributes. In practice, the relevance of numerical attributes is highly dependent on the relational context. For instance, given the query (\textit{San Francisco}, \textit{Less\_Elders}, \textit{x}), the attribute \textit{median\_age} becomes critical; whereas in (\textit{San Francisco}, \textit{Easier\_Commuting}, \textit{x}), the key attribute is \textit{commute\_time}. Models lacking context-aware alignment treat numerical features uniformly, leading to semantic incompleteness and degraded reasoning performance. (see Figure~\ref{NBA-KG}).

    \item[\textbf{C2}] \textbf{Ordinal indistinguishability.} Numerical reasoning often requires fine-grained distinction between close values. However, existing models typically learn coarse semantic representations of numerical attributes, making it difficult to capture such subtle ordinal relations. This leads to confusion in tasks that rely on comparative inference, and the ambiguity becomes more pronounced when handling hard negative samples with near-equal values.
\end{enumerate}

To address these challenges, we propose \textbf{NumCoKE}, a novel framework for numerical reasoning on knowledge graphs via Mixture-of-Experts and Ordinal Contrastive Embedding. To tackle (\textbf{C1)}, we design a \textit{Mixture-of-Experts Knowledge-Aware} (MoEKA) encoder that dynamically captures the contextual importance of numeric attributes under different relational settings. Specifically, we construct relation-aware expert modules that encode entities and relations into a shared vector space. Through adaptive routing, numerical attributes are directed to the most relevant experts based on their semantic context. This mechanism enables NumCoKE to selectively amplify semantically important attributes for each relation-entity pair, thereby producing fine-grained and context-sensitive representations.
To tackle (\textbf{C2}), we propose an \textit{Ordinal Knowledge Contrastive Learning} (OKCL) strategy. Instead of relying solely on binary positive-negative sampling, OKCL synthesizes ordinal sample triplets by computing cosine-based similarity among value distributions. We then select top-$k$ ordinal neighbors to serve as fine-grained supervision signals. This strategy preserves the continuity of numerical features and explicitly encourages the model to recognize nuanced ordinal relations between close values (e.g., distinguishing \textit{183cm} from \textit{184cm}). We further provide a theoretical analysis to justify the consistency and effectiveness 
of OKCL.

We validate the effectiveness of NumCoKE through extensive experiments on three real-world benchmark datasets covering diverse types of numerical attribute distributions (discrete, continuous, and skewed). The results show that our method consistently outperforms strong baselines across all settings. In summary, our contributions are three-folds:
\begin{itemize}
\item We propose \textbf{NumCoKE}, a novel and efficient model for numerical reasoning over KGs, which for the first time incorporates relation-aware mixture-of-experts encoding to dynamically integrate entity, relation, and attribute semantics.
\item We introduce ordinal knowledge contrastive learning, a new contrastive paradigm that generates ordinally structured samples to better capture fine-grained numerical distinctions, improving the model’s sensitivity to subtle ordinal relationships.
\item We conduct comprehensive experiments on three public datasets. NumCoKE achieves state-of-the-art performance, demonstrating superior accuracy of our model.
\end{itemize}

\section{Related Work}
\noindent\textbf{Knowledge Graph Embedding (KGE).}
KGE amis to capture the latent representations of entities and relations in KGs. TransE \cite{DBLP:conf/nips/BordesUGWY13} and its extended models \cite{ DBLP:journals/jifs/ZhangJWZKQL21, DBLP:conf/coling/LiSZG24}, focus on treating relation as a \textit{translation} from the head entity to the tail entity.
DaBR \cite{DBLP:conf/coling/WangLBG25} utilize distance-adaptive translations to learn geometric distance between entities,
KGDM \cite{DBLP:conf/aaai/LongZLWLW24} estimates the probabilistic distribution of target entities in prediction through diffusion models.
RuIE \cite{DBLP:conf/coling/LiaoDH025} leverage logical rules to enhance reasoning.
ConvE \cite{DBLP:conf/aaai/DettmersMS018} extracts deep features of head entity and relation based on 2D convolution. ConE \cite{DBLP:conf/nips/BaiYRL21}, MuRP \cite{DBLP:conf/nips/BalazevicAH19}, GIE \cite{DBLP:conf/aaai/CaoX0CH22}, AttH \cite{DBLP:conf/acl/ChamiWJSRR20} and LorentzKG \cite{DBLP:conf/acl/Fan0CCDY24} embed KGs into hyperbolic spaces to model hierarchical relations. Further, models \cite{DBLP:conf/esws/SchlichtkrullKB18, DBLP:conf/aaai/ShangTHBHZ19, DBLP:conf/iclr/VashishthSNT20} of GNNs have been proposed to model higher-order connectivity in knowledge graphs.
However, these models do not consider numerical values, making it difficult to accomplish numerical reasoning tasks.

\noindent\textbf{KGE with Numerical Attributes.}
To improve the performance of KGE, several KGE models have been proposed to incorporate auxiliary information about the entity such as literals and numerical attributes.
For example, KBLRN \cite{DBLP:conf/uai/Garcia-DuranN18} accomplish the KGE task based on numerical values, considering numerical differences between different entities. MT-KGNN \cite{DBLP:conf/cikm/TayTPH17} is a multitask model to predict both numeric attribute values and entity/relation embeddings. LiteralE \cite{DBLP:conf/semweb/KristiadiKL0F19} combines the entity embedding with numeric attributes based on a learnable gating function.
Deterministic stand-alone value representation methods including NEKG \cite{DBLP:conf/emnlp/DuanYT21} and NRN \cite{DBLP:conf/kdd/BaiLLYYS23} are also used to predict numerical attributes in KGs. However, these models do not exploit joint interactions among entities, relations and numerical attributes to mine the useful semantic information.


\noindent\textbf{Contrastive Learning (CL).}
CL is a self-supervised learning method by pulling semantically close neighbors together while pushing non-neighbors away, which can improving the representations of entities/graphs. SimGCL \cite{DBLP:conf/sigir/YuY00CN22} adds uniform noise to the entity representation, which is an expansion-free CL method. HeCO \cite{DBLP:conf/kdd/WangLHS21} employs contrast learning in network schema and meta-path, capturing both local and global features of entities. SLiCE \cite{DBLP:conf/www/WangAHCR21} makes use of mutual attraction between closed nodes to learn subgraph representations. RAKGE \cite{DBLP:conf/kdd/KimKKPJP23} interacts with training samples with head entities to generate samples. RRNE \cite{DBLP:conf/pakdd/JeongJKKKP25} selects hard samples from subgraph. However, none of the above methods can guarantee effective samples for numerical reasoning tasks.


\section{Methodology}
\begin{figure*}[htbp]
    \centering
    \includegraphics[scale=0.5]{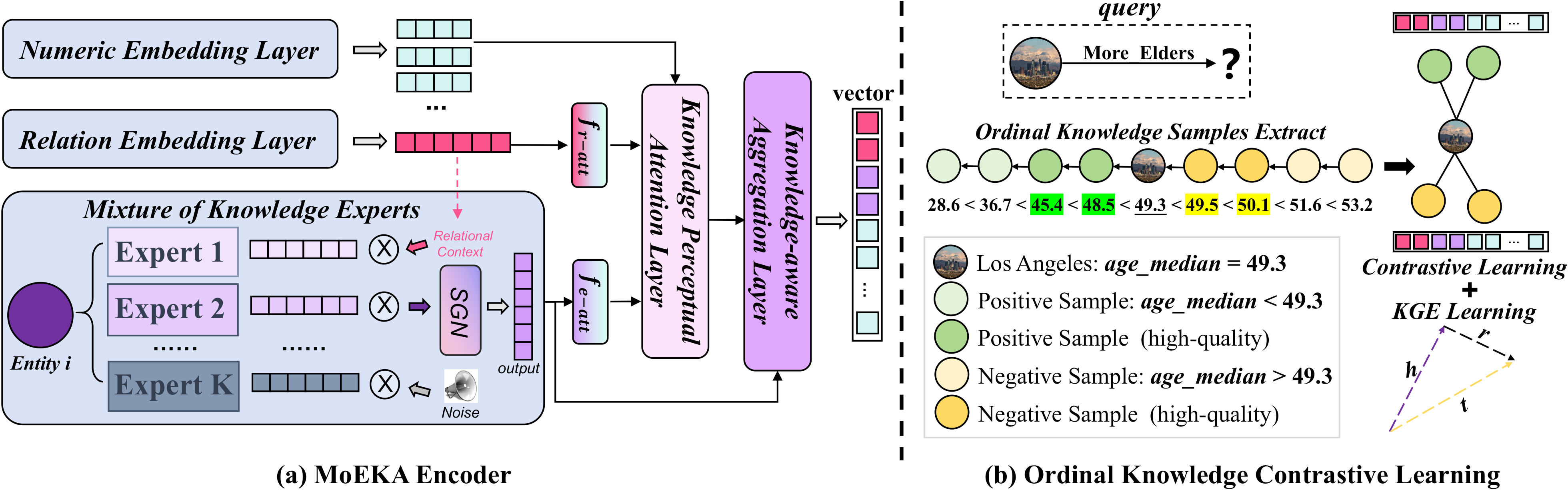}
    \caption{
   The Overview of our model. 
    (a) The MoEKA Encoder encodes each entity with the relation and attributes to a unified, elaborate semantic representation. 
    (b) To capture the fine-grained semantic information, we utilize a new knowledge
contrastive learning method to generate high-quality ordinal samples to learn the nuances in attributes and distinguish similar semantics.
    }
    \label{J-ERA}
\end{figure*}

In this section, we present a new numerical reasoning framework for KG termed NumCoKE.
Specifically, there are two main components: Mixture-of-Experts-Knowledge-Aware Encoder (MoEKA-Encoder) and ordinal knowledge contrastive learning (OKCL) strategy.
The overall framework of NumCoKE is illustrated in Figure \ref{J-ERA}.

\noindent\textbf{Problem Definition.}
Given a knowledge graph as a collection of real triples $\mathcal{G} = \{(h,r,t)\}$, where  $h,t\in\mathcal{E}$ and $r\in\mathcal{R}$ represent the set of entities and relations, respectively. 
Denote an entity-numeric value matrix for entities as $\bm{X}\in \mathbb{R}^{\vert\mathcal{E}\vert \times \vert\mathcal{M}\vert}$ and the \textit{m}-th numeric value belongs to the entity \textit{i} as $X_{im}$, where $\mathcal{M}$ is the set of numeric attribute fields (e.g., \textit{height}, \textit{weight}, and \textit{born\_{}year}). Due to the incompleteness of the KG, KGE emerge as an essential approach to conduct the knowledge graph completion, which is to map each triple (\textit{h, r, t}) to a reasonableness score based on the KGE model, where a low score means that the triple is not reasonable, and vice versa. In this paper, our purpose is to construct a new KGE model to conduct numerical reasoning with the assistance of numerical attributions. 

\subsection{Knowledge-Aware Learning}

\noindent\textbf{Numeric Value Embedding Learning.}
To learn the numeric values, for observable scalars, we use a learnable embedding matrix to map them to a vector space. For missing scalars, we use learnable special missing value embeddings to preserve the magnitude information and prevent the task from being threatened, instead of using fixed values such as zero: $\bm{o}_{i}^{m} = (\bm{W}_{x}^{m} + \bm{w}_{m}X_{im}) \odot \bm{v}_{m}$,
where $X_{im}$ is the corresponding numeric value of the entity \textit{i}. $\bm{v}_{m} \in \mathbb{R}^{d_{att}}$ represents the embedding vector of the numeric attribute field \textit{m}, $d_{att}$ stands for the embedding dimension of attribute. $\bm{w}_{m} \in \mathbb{R}^{d_{att}}$ and $\bm{W}_{x}^{m} \in \mathbb{R}^{d_{att} \times d_{att}}$ are linear transformations that learn the context information between each numeric value and attribute. The operation symbol $\odot$ represents point-wise multiplication. We denote $\bm{o}_{i}^{m} \in \mathbb{R}^{d_{att}}$ as the \textit{m}-th field attribute embedding of the entity \textit{i}.
In this way, we can learn a more robust representation of both existing/missing numerical attributes.

\noindent\textbf{Mixture of Knowledge Experts (MOE).} 
To better learn entity embeddings from different perspectives with relation contexts and numerical attributes, we first employ a module called Mixture of Knowledge Experts to build expert networks, where each perspective corresponds to an expert network. First, the entity $i \in\mathcal{E}$ has a raw feature $\tilde{e}_{i}$. We then learn the multi-perspective embeddings $\mathcal{H}^{\tilde{e}_{i}}_{1}, \mathcal{H}^{\tilde{e}_{i}}_{2}, \ldots, \mathcal{H}^{\tilde{e}_{i}}_{K}$ for the entity $i$ by establishing $K$ knowledge experts denoted as $\mathcal{W}_{1}, \mathcal{W}_{2}, \ldots, \mathcal{W}_{K}$. This process can be represented as $\mathcal{H}^{\tilde{e}_{i}}_{k} = \mathcal{W}_{k}(\tilde{e}_{i})$. Next, we design a semantic-guided gated fusion network (SGN) to facilitate the fusion of inter-perspective entity embeddings with relation guidance.
\begin{equation}
    \bm{e}_{i} = \sum\nolimits_{k=1}^{K}\mathcal{F}_{k}(\mathcal{H}^{\tilde{e}_{i}}_{k},r)\mathcal{H}^{\tilde{e}_{i}}_{k}
\end{equation} 
where $\bm{e_{i}}\in \mathbb{R}^{d_{emb}}$ is the output entity embedding of entity \textit{i} for relation $r$, $\mathcal{F}_{k}$ is the weight for each expert: 
\begin{equation}
    \mathcal{F}_{k}(\mathcal{H}^{\tilde{e}_{i}}_{k}, r) = \frac{exp((\mathcal{U}(\mathcal{H}^{\tilde{e}_{i}}_{k})+\psi_{k})/\rho(\epsilon_{r}))}
    {\sum\nolimits_{j=1}^{K}exp((\mathcal{U}(\mathcal{H}^{\tilde{e}_{i}}_{j})+\psi_{j})/\rho(\epsilon_{r}))}
\end{equation}
where $\psi_{k} \sim \mathcal{N}(0, \mathcal{U'}(\mathcal{H}^{\tilde{e}_{i}}_{k}))$, $\mathcal{U}$ and $\mathcal{U' }$ are two projection layers, and $\psi_{k}$ is tunable Gaussian noise \cite{DBLP:conf/cikm/BianPZWWW23} used to balance the weights for each expert and enhance the model's robustness. Additionally, we add a relation-aware temperature $\epsilon_{r}$ with a sigmoid function $\rho$ to limit the temperature within the range $(0,1)$. Our aim is to obtain an entity embedding within the relational context of the current prediction before making the final decision. 
\noindent\textbf{Knowledge Perceptual Attention.}
 Based on the representations above,  we then learn the knowledge-aware representations via the joint semantic interactions among entities, relations, and attributes. In this way, we propose a \textit{knowledge perceptual attention mechanism} for entity-relation attributes. Specifically, we leverage a single-layer perception $f_{e-att}$ and $f_{r-att}$ to project the entity and relation embeddings onto the attribute embedding subspace, respectively. Thereafter, we mixing the information of entity \textit{e} and relation \textit{r}, so as to get the joint projected embedding $\bm{p}_{joint}^{att}\in\mathbb{R}^{d_{att}}$:
\begin{equation}
    \bm{p}_{joint}^{att} = \sum_{*\in\{e, r\}} \delta_{*}f_{*-att}(\bm{p}_{*})
    \label{EQ Attention}
\end{equation}
where $\bm{p}_{e}$ corresponds to $\bm{e}_{i}$, $\bm{p}_{r}$ corresponds to the relation embedding, $\delta_{*} \in [0,1]$ are hyperparameters and $\delta_{e} + \delta_{r} = 1$. 

Then, we apply the \textit{knowledge perceptual attention mechanism} to establish the joint interactions between entities, relations, and attributes.
Specifically, We apply this mechanism for all attribute fields $m$ = 1,2,$\cdots,\vert\mathcal{M}\vert$ and we repeat this step for all attention heads
\textit{l} = 1,2,$\cdots$,$L$, where $L$ is the number of multi-head attentions. The result of the joint interaction of head \textit{l} is shown as follows:
\begin{equation}
    \bm{o}_{joint,i}^{(l)} = \sum\nolimits_{m=1}^{| \mathcal{M}|}a_{joint,i,m}^{(l)}(\bm{W}_{agg}^{(l)}\bm{o}_{i}^{m})
    \label{EQ9}
\end{equation}
\noindent where the linear transformation matrix $\bm{W}_{agg}^{(l)} \in \mathbb{R}^{d_{sub} \times d_{att}}$ aims to project attribute embedding $\bm{o}_{i}^{m}$ into low-dimensional subspaces to capture the importance of each numeric attribute from the given entity with relation accurately. Under a specific attention head \textit{l}, we map $\bm{e}_{joint}^{att}$ and $\bm{o}_{i}^{1}, \cdots \bm{o}_{i}^{\vert\mathcal{M}\vert} \in \mathbb{R}^{d_{att}}$ onto smaller spaces and capture the attention score. The normalized attention weight $a_{joint,i,m}^{(l)}$ of the attribute embedding $\bm{o}_{i}^{m}$ is formulated as follows:
\begin{equation}
    a_{joint,i,m}^{(l)} = \frac{exp(s^{(l)}(\bm{p}_{joint}^{att},\bm{o}_{i}^{m}))}{\sum\nolimits_{n=1}^{\vert \mathcal{M}\vert}exp(s^{(l)}(\bm{p}_{joint}^{att},\bm{o}_{i}^{n}))}
    \label{EQ8}
\end{equation}
\begin{equation}
s^{(l)}(\bm{p}_{joint}^{att},\bm{o}_{i}^{m}) = \sigma \left(\|_{j=1}^H  \bm{W}_j^{(l)} \phi_j(\bm{p}_{joint}^{att},\bm{o}_{i}^{m}) \right)
\end{equation}
 \normalsize
where $\sigma$ denotes the LeakyReLU activation function,  $\bm{W}_j^{(l)}$ denotes the weighted matrix corresponding to the $i$-th operator in the  $l$-th layer,  $\phi_j((\bm{p}_{joint}^{att},\bm{o}_{i}^{m}))$ denotes the fusion operation, which can be model by MLP. $\|$ here denotes the multi-head mechanism.

\noindent\textbf{Knowledge-aware Aggregation.} The final attribute embedding for entity \textit{i} related to the joint perceptual vector is as follows:
$\bm{o}_{joint,i} =  \bm{o}_{joint,i}^{(1)} \|  \bm{o}_{joint,i}^{(2)} \|
\cdots \|
 \bm{o}_{joint,i}^{(L)}$,
where $\bm{o}_{joint,i} \in \mathbb{R}^{d_{att}}$ as the attribute vector of entity $i$, which is aware of the entity and the relation. Symbol $\|$ represents a concatenation operator. To balance the joint knowledge-aware attribute vectors and the embedded entities, we have:
\begin{equation}
    \bm{e}_{joint,i}^{att} = \sigma(\bm{W}_{ge}\bm{e}_{i}+\bm{W}_{ga}\bm{o}_{joint,i})+\bm{b}
    \label{GRU}
\end{equation}
where 
$\bm{W_}{ge} \in \mathbb{R}^{d_{emb} \times d_{emb}}$ and 
$\bm{W}_{ga} \in \mathbb{R}^{d_{emb} \times d_{att}}$ are different linear transformations to assign different weights to each entity representation and fused numerical representation by adaptively learning to obtain high quality final embeddings and to focus on the more important information. $\bm{b}$ is a bias.
$\sigma$  is a sigmoid function. We name $\bm{e}_{joint,i}^{att} \in \mathbb{R}^{d_{emb}}$ as the attribute-enriched vector of entity \textit{i}.

\subsection{Ordinal Knowledge Contrastive Learning}
Learning ordinal relations is essential for numerical reasoning tasks. 
Unlike the previous work \cite{DBLP:conf/kdd/KimKKPJP23} mainly focusing on generating valid, positive, and negative samples, we turn to a more difficult and significant task that generates high-quality ordinal samples. 

\noindent\textbf{Ordinal Relation Learning}. Given \( N \) objects and the ordinal relationship triples $\mathcal{S} = \{ (a, b, c) \mid a, b, c \in \mathcal{E}, a \neq b \neq c \}$, where \([N] = \{1, \ldots, N\}\) and object \( a \) is more similar to object \( b \) than it is to object \( c \). Given some distance function \( d(\cdot, \cdot) \), we aim to learn the representations of objects, denoted as \( \{\mathbf{x}_1, \mathbf{x}_2, \ldots, \mathbf{x}_N\} \), such that the following objectives hold as much as possible:
\begin{equation}\label{distance of OE3}
 d(\mathbf{x}_a, \mathbf{x}_b) < d(\mathbf{x}_a, \mathbf{x}_c), \quad \forall (a, b, c) \in \mathcal{S}    
\end{equation}

\noindent\textbf{Knowledge Samples Generator.} On the analysis above, to distinguish precise semantics in numerical reasoning, we generate the preliminary positive/negative ($\mathcal{E}^+$ and $\mathcal{E}^-$) samples based on the available head entities and relations, the set of positive/negative samples is as follows:
\begin{equation}
    \centering
    \mathcal{E^{+}} = \{
    \bm{e}_{joint,i}^{att}\vert i \in \mathcal{P}[h,r]\},\ 
    \centering
    \mathcal{E^{-}} = \{ \bm{e}_{joint,j}^{att}\vert j \in \mathcal{N}[h,r]\}
    \label{EQ15}
\end{equation}
where $\mathcal{P}[h,r]$ represents the set of positive tail entities, 
$\mathcal{N}[h,r]$ represents the set of positive tail entities. 

\noindent\textbf{Ordinal Samples Extractor.} Since each attribute field (e.g., \textit{year}, \textit{age}, \textit{weight}) has its unique distribution, we select k number of samples with the highest cosine-similarity to $\bm{e}_{joint,i}^{att}$ from all the generated positive/negative samples with the assistance of Eq. (\ref{distance of OE3}). Note that this step is essential and it can generate the samples obeying ordinal embedding. Then we generate high-quality ordinal samples as follows: 
\begin{equation}
    \bm{e}_{mix}^{t+} = \alpha \cdot \sum_{p \in Topk(\mathcal{E^{+}})}\bm{e}_{joint,p}^{att} + (1 - \alpha) \cdot \bm{e}_{joint,h}^{att}
    \label{sample positive}
\end{equation}
\begin{equation}
    \bm{e}_{mix}^{t-} = \beta \cdot \sum_{n \in 
    Topk(\mathcal{E^{-}})}\bm{e}_{joint,n}^{att} + (1 - \beta) \cdot \bm{e}_{joint,h}^{att}
    \label{sample negative}
\end{equation}
where $\alpha,\beta$ are blending coefficients that are sampled from the uniform distribution $[0,1]$. $Topk(\mathcal{E^{+}})$ and $Topk(\mathcal{E^{-}})$ are sampling sets with the highest cosine similarity to $\bm{e}_{joint, i}^{att}$ in $\mathcal{E}^+$ and $\mathcal{E}^-$, respectively. 
In this sense, to satisfy ordinal relation, we have:
\begin{equation}
    d(\bm{e}_{joint,h}^{att},p)<d(\bm{e}_{joint,h}^{att},n)
\end{equation}
where $p\in Topk(\mathcal{E^{+}})$ and $n\in Topk(\mathcal{E^{-}})$. In this way,
 our contrastive learning loss is defined as follows:
\begin{equation}
\small
\label{EQ20}
    \begin{split}
        \mathcal{L}_{CL} = - & \frac{1}{\mathcal{|G|}}\sum\nolimits_{l \in \mathcal{G}}\log \frac{\varpi(\textbf{p}^{\top}\textbf{u}/\tau)}{\varpi(\textbf{p}^{\top}\textbf{u}/\tau)+\sum\nolimits_{v_{-} \in \bm{e}_{mix}^{t-}}\varpi(\textbf{v}^{\top}\textbf{v}_{-}/\tau)}
    \end{split}
\end{equation}
where $\bm{p}=\bm{e}_{joint,h}^{att}+\bm{e}_r$, $\bm{u}=\bm{e}_{mix}^{t+}$, 
$\tau$ is the temperature hyperparameter. $\varpi$ represents exponential function $exp(\cdot$). By implementing the strategy of randomly mixing the positive tails and employing head blending techniques (as described in Eqs. (\ref{sample positive})-(\ref{sample negative})), we notably increase the diversity of our samples and interpret the nuances of semantic information.

\subsection{Training}
\noindent\textbf{Score Function.}
A classical assumption of typical KGE methods is to compute the distances between head entities, relations, and tail entities. However, in numerical reasoning tasks, these methods suffer from serious drawbacks because they do not effectively address the order embedding problem posed by relational patterns in complex tasks. For example, given the head entity (\textit{weight=80kg}) and its relation (\textit{is\_{}heavier\_{}than}), and using the TransE score function, the two tail entities $e_{1}$ (\textit{weight=70kg}) and $e_{2}$ (\textit{weight=60kg}) are mapped to the same location, failing to express a numerical ordering relation such as 80\textgreater70\textgreater60, which impedes numerical reasoning. To address this issue, we design a rationality score function specifically for the numerical reasoning task as follows:
\begin{equation}
    \begin{split}
&score(\bm{e}_{joint,h}^{att},\bm{e}_{r},\bm{e}_{joint,i}^{att}  )\\ = & \varepsilon -  \Vert \bm{e}_{joint,h}^{att}+\bm{e}_{r}-\bm{e}_{joint,i}^{att} \Vert_{1/2} + \\& \Delta \Vert max(0,  \bm{W}_{r}\bm{e}_{joint,h} - \bm{W}_{r}\bm{e}_{joint,i}^{att}) \Vert^{2}
    \end{split}
    \label{EQ21}
\end{equation}
where $\Delta$ represents the weight in the numerical reasoning score function. The first term is derived from TransE \cite{DBLP:conf/nips/BordesUGWY13},  
while the second term modifies the Order-Embedding \cite{DBLP:journals/corr/VendrovKFU15}.
$\bm{W}_{r} \in \mathbb{R}^{d_{emb} \times d_{emb}}$ is a projection matrix critical for mapping entities to a specific relational space, indicating that depending on the type of relation, each entity may assume a different order.

\noindent\textbf{Loss Function.}
In this paper, we use the binary cross-entropy loss \cite{DBLP:conf/aaai/DettmersMS018,DBLP:conf/semweb/KristiadiKL0F19} to optimize the training process. Let $\mathcal{T} = \mathcal{G} \cup \mathcal{G}^{-}$ denote the
training dataset,where $\mathcal{G}$ denotes the set of positive knowledge triples, $\mathcal{G}^{-}$ denotes the set of negative knowledge triples 
$\{(h,r,t^{'})|h,t^{'} \in \mathcal{E},r \in \mathcal{R},(h,r,t^{'}) \notin \mathcal{G} \}$. The binary-cross entropy loss is defined as follows:
\begin{equation}
    \mathcal{L}_{BCE} = - \frac{1}{\vert\mathcal{T}\vert} \sum\limits_{l \in \mathcal{T}}(y_{l} \log(p_{l}) + (1 - y_{l})\log(1-p_{l})) 
    \label{EQ22}
   \end{equation}
\noindent where  $y_{l} \in \{0,1\}$ is the truth label, and $p_{l} \in$ [0,1] is the probability of each triple  $(h,r,t)=l\in \mathcal{T}$, which is formulated as Eq. (\ref{EQ21}). Combining Eq. (\ref{EQ20}) and Eq. (\ref{EQ22}), the final loss can be summarized as follows:
\begin{equation}
    \mathcal{L}_{total} = \mathcal{L}_{BCE} + \lambda \mathcal{L}_{CL}
    \label{EQ23}
\end{equation}
\noindent where $\lambda$ is a hyperparameter, stands for the coefficient of contrastive learning loss.


\begin{table*}[t]
\centering
\scriptsize
\setlength{\tabcolsep}{4pt}
\begin{tabular}{l|l|lllll|lllll|lllll}
\hline
\multicolumn{2}{c|}{\multirow{2}{*}{\textbf{Model}}}& \multicolumn{5}{|c|}{\textbf{US-Cities}}      & \multicolumn{5}{c|}{\textbf{Spotify}}& \multicolumn{5}{c}{\textbf{Credit}}\\
\cline{3-17}
\multicolumn{2}{c|}{\quad}& \textbf{H@1}& \textbf{H@3}&\textbf{H@10} &\textbf{MR}& \textbf{MRR}& \textbf{H@1} & \textbf{H@3}&\textbf{H@10}& \textbf{MR}& \textbf{MRR} & \textbf{H@1}& \textbf{H@3}&\textbf{H@10} & \textbf{MR} & \textbf{MRR}\\
\hline
\multirow{6}{*}{\textbf{\textit{Euclidean}}}&\textbf{TransE}&\cellcolor{pink!30}0.189&\cellcolor{pink!30}0.248&\cellcolor{pink!30}0.324&\cellcolor{pink!30}367&\cellcolor{pink!30}0.239&\cellcolor{pink!30}0.259&\cellcolor{pink!30}0.355&\cellcolor{pink!30}0.462&\cellcolor{pink!30}115&\cellcolor{pink!30}0.332&\cellcolor{pink!30}0.421&\cellcolor{pink!30}0.520&\cellcolor{pink!30}0.630&\cellcolor{pink!30}39&\cellcolor{pink!30}0.493 \\
& \textbf{ConvE}&\cellcolor{pink!30}0.158&\cellcolor{pink!30}0.202&\cellcolor{pink!30}0.274&\cellcolor{pink!30}385&\cellcolor{pink!30}0.198&\cellcolor{pink!30}0.231&\cellcolor{pink!30}0.307&\cellcolor{pink!30}0.414&\cellcolor{pink!30}110&\cellcolor{pink!30}0.295&\cellcolor{pink!15}0.171&\cellcolor{pink!30}0.281&\cellcolor{pink!30}0.430&\cellcolor{pink!15}58&\cellcolor{pink!15}0.261 \\
& \textbf{TuckER}&\cellcolor{pink!30}0.156&\cellcolor{pink!30}0.212&\cellcolor{pink!30}0.308&\cellcolor{pink!30}321&\cellcolor{pink!30}0.207&\cellcolor{pink!30}0.211&\cellcolor{pink!30}0.296&\cellcolor{pink!30}0.411&\cellcolor{pink!30}98&\cellcolor{pink!30}0.278&\cellcolor{pink!30}0.405&\cellcolor{pink!30}0.512&\cellcolor{pink!30}0.638&\cellcolor{pink!30}36&\cellcolor{pink!30}0.485 \\
& \textbf{HAKE}&\cellcolor{pink!5}0.003&\cellcolor{pink!5}0.025&\cellcolor{pink!5}0.064&\cellcolor{pink!5}981&\cellcolor{pink!5}0.024&\cellcolor{pink!5}0.008&\cellcolor{pink!15}0.085&\cellcolor{pink!15}0.120&\cellcolor{pink!15}196&\cellcolor{pink!5}0.075&\cellcolor{pink!15}0.051&\cellcolor{pink!15}0.151&\cellcolor{pink!15}0.273&\cellcolor{pink!15}128&\cellcolor{pink!15}0.134 \\
&
\textbf{DaBR}&\cellcolor{pink!35}0.202&\cellcolor{pink!30}0.256&\cellcolor{pink!30}0.337&\cellcolor{pink!30}352&\cellcolor{pink!30}0.241&\cellcolor{pink!30}0.266&\cellcolor{pink!35}0.367&\cellcolor{pink!30}0.477&\cellcolor{pink!30}104&\cellcolor{pink!30}0.334&\cellcolor{pink!30}0.425&\cellcolor{pink!30}0.526&\cellcolor{pink!30}0.621&\cellcolor{pink!30}36&\cellcolor{pink!30}0.498\\
& \textbf{KGDM}&\cellcolor{pink!35}0.232&\cellcolor{pink!35}0.271&\cellcolor{pink!30}0.353&\cellcolor{pink!30}329&\cellcolor{pink!35}0.246&\cellcolor{pink!30}0.274&\cellcolor{pink!35}0.374&\cellcolor{pink!30}0.489&\cellcolor{pink!35}99&\cellcolor{pink!30}0.345&\cellcolor{pink!35}0.419&\cellcolor{pink!30}0.532&\cellcolor{pink!30}0.626&\cellcolor{pink!30}39&\cellcolor{pink!30}0.482\\
& 
\textbf{RuIE}&\cellcolor{pink!15}0.084&\cellcolor{pink!15}0.126&\cellcolor{pink!15}0.185&\cellcolor{pink!15}587&\cellcolor{pink!15}0.120&\cellcolor{pink!15}0.124&\cellcolor{pink!15}0.188&\cellcolor{pink!15}0.284&\cellcolor{pink!15}173&\cellcolor{pink!15}0.182&\cellcolor{pink!30}0.291&\cellcolor{pink!30}0.388&\cellcolor{pink!30}0.513&\cellcolor{pink!15}65&\cellcolor{pink!15}0.366 \\

\hline
\multirow{5}{*}{\textbf{\textit{Hyperbolic}}}

&\textbf{AttH}&\cellcolor{pink!15}0.051&\cellcolor{pink!15}0.069&\cellcolor{pink!15}0.108&\cellcolor{pink!15}1255&\cellcolor{pink!15}0.076 &\cellcolor{pink!15}0.052&\cellcolor{pink!15}0.082&\cellcolor{pink!15}0.146&\cellcolor{pink!5}348&\cellcolor{pink!5}0.087&\cellcolor{pink!15}0.176&\cellcolor{pink!15}0.261&\cellcolor{pink!15}0.395&\cellcolor{pink!15}101&\cellcolor{pink!15}0.251\\

&\textbf{ConE}&\cellcolor{pink!5}0.009&\cellcolor{pink!5}0.042&\cellcolor{pink!15}0.114&\cellcolor{pink!30}239&\cellcolor{pink!5}0.048&\cellcolor{pink!5}0.001&\cellcolor{pink!5}0.006&\cellcolor{pink!5}0.064&\cellcolor{pink!5}227&\cellcolor{pink!5}0.028&\cellcolor{pink!5}0.006&\cellcolor{pink!15}0.241&\cellcolor{pink!15}0.395&\cellcolor{pink!15}82&\cellcolor{pink!15}0.156\\

&\textbf{GIE}&\cellcolor{pink!15}0.095&\cellcolor{pink!15}0.134&\cellcolor{pink!15}0.206&\cellcolor{pink!15}570&\cellcolor{pink!15}0.132&\cellcolor{pink!15}0.121&\cellcolor{pink!15}0.183&\cellcolor{pink!15}0.281&\cellcolor{pink!15}185&\cellcolor{pink!15}0.174&\cellcolor{pink!30}0.285&\cellcolor{pink!15}0.388&\cellcolor{pink!15}0.512&\cellcolor{pink!15}76&\cellcolor{pink!15}0.359\\

&\textbf{MuRP}&\cellcolor{pink!15}0.082&\cellcolor{pink!15}0.113&\cellcolor{pink!15}0.175&\cellcolor{pink!15}457&\cellcolor{pink!15}0.115&\cellcolor{pink!15}0.024&\cellcolor{pink!15}0.179&\cellcolor{pink!15}0.324&\cellcolor{pink!15}135&\cellcolor{pink!15}0.119&\cellcolor{pink!15}0.151&\cellcolor{pink!15}0.246&\cellcolor{pink!15}0.432&\cellcolor{pink!15}80&\cellcolor{pink!15}0.228\\

&\textbf{LorentzKG}&\cellcolor{pink!30}0.179&\cellcolor{pink!30}0.221 &\cellcolor{pink!30}0.286 &\cellcolor{pink!30}368  &\cellcolor{pink!30}0.209 &\cellcolor{pink!30}0.239&\cellcolor{pink!30}0.312&\cellcolor{pink!30}0.412&\cellcolor{pink!30}126&\cellcolor{pink!30}0.320 &\cellcolor{pink!30}0.402&\cellcolor{pink!30}0.509&\cellcolor{pink!30}0.597&\cellcolor{pink!30}38&\cellcolor{pink!30}0.482\\

\hline
\multirow{2}{*}{\textbf{\textit{GNNs}}}&\textbf{R-GCN}&\cellcolor{pink!30}0.212&\cellcolor{pink!30}0.271&\cellcolor{pink!30}0.355&\cellcolor{pink!30}314&\cellcolor{pink!30}0.263&\cellcolor{pink!45}0.288&\cellcolor{pink!45}0.382&\cellcolor{pink!45}0.504&\cellcolor{pink!30}89&\cellcolor{pink!45}0.364&\cellcolor{pink!45}0.480&\cellcolor{pink!45}0.570&\cellcolor{pink!45}0.675&\cellcolor{pink!30}34&\cellcolor{pink!45}0.546\\
& \textbf{WGCN}&\cellcolor{pink!15}0.029&\cellcolor{pink!5}0.056&\cellcolor{pink!15}0.104&\cellcolor{pink!5}968&\cellcolor{pink!5}0.057&\cellcolor{pink!15}0.095&\cellcolor{pink!15}0.162&\cellcolor{pink!15}0.261&\cellcolor{pink!15}331&\cellcolor{pink!15}0.153&\cellcolor{pink!15}0.170&\cellcolor{pink!15}0.256&\cellcolor{pink!15}0.369&\cellcolor{pink!15}100&\cellcolor{pink!15}0.241\\
\hline
\multirow{3}{*}{\textbf{\textit{Attributed}}}

&\textbf{KBLRN}&\cellcolor{pink!5}0.006&\cellcolor{pink!5}0.018&\cellcolor{pink!5}0.046&\cellcolor{pink!5}2164&\cellcolor{pink!5}0.021&\cellcolor{pink!5}0.017&\cellcolor{pink!5}0.039&\cellcolor{pink!5}0.086&\cellcolor{pink!5}347&\cellcolor{pink!5}0.044&\cellcolor{pink!5}0.007&\cellcolor{pink!5}0.020&\cellcolor{pink!5}0.079&\cellcolor{pink!5}271&\cellcolor{pink!5}0.061\\

& \textbf{MT-KGNN}&\cellcolor{pink!15}0.071&\cellcolor{pink!15}0.109&\cellcolor{pink!15}0.156&\cellcolor{pink!15}653&\cellcolor{pink!15}0.102&\cellcolor{pink!15}0.108&\cellcolor{pink!15}0.182&\cellcolor{pink!15}0.302&\cellcolor{pink!15}142&\cellcolor{pink!15}0.176&\cellcolor{pink!15}0.211&\cellcolor{pink!15}0.304&\cellcolor{pink!15}0.436&\cellcolor{pink!15}74&\cellcolor{pink!15}0.285\\

&\textbf{LiteralE}&\cellcolor{pink!45}0.246&\cellcolor{pink!45}0.308&\cellcolor{pink!45}0.402&\cellcolor{pink!45}228&\cellcolor{pink!45}0.299&\cellcolor{pink!45}0.264&\cellcolor{pink!45}0.371&\cellcolor{pink!45}0.498&\cellcolor{pink!45}76&\cellcolor{pink!45}0.345&\cellcolor{pink!45}0.475&\cellcolor{pink!45}0.562 &\cellcolor{pink!45}0.676 &\cellcolor{pink!30}36  &\cellcolor{pink!45}0.549 \\

\hline
\multirow{2}{*}{\textbf{\textit{Deterministic}}}
& \textbf{NEKG}&\cellcolor{pink!30}0.196&\cellcolor{pink!30}0.230 &\cellcolor{pink!30}0.304 &\cellcolor{pink!30}376  &\cellcolor{pink!30}0.217 &\cellcolor{pink!30}0.246&\cellcolor{pink!30}0.323&\cellcolor{pink!30}0.423&\cellcolor{pink!30}121&\cellcolor{pink!30}0.336 &\cellcolor{pink!30}0.398&\cellcolor{pink!30}0.512&\cellcolor{pink!30}0.589&\cellcolor{pink!30}41&\cellcolor{pink!30}0.477\\
& 
\textbf{NRN}&\cellcolor{pink!45}0.298&\cellcolor{pink!45}0.326&\cellcolor{pink!45}0.432&\cellcolor{pink!60}155&\cellcolor{pink!45}0.323&\cellcolor{pink!45}0.342&\cellcolor{pink!45}0.442&\cellcolor{pink!45}0.546&\cellcolor{pink!45}102&\cellcolor{pink!45}0.414&\cellcolor{pink!45}0.560&\cellcolor{pink!45}0.639&\cellcolor{pink!45}0.740&\cellcolor{pink!45}21&\cellcolor{pink!45}0.620\\
\hline
\multirow{4}{*}{\textbf{\textit{Self-supervised}}} &\textbf{BiGI}&\cellcolor{pink!30}0.185&\cellcolor{pink!30}0.249&\cellcolor{pink!30}0.331&\cellcolor{pink!30}359&\cellcolor{pink!30}0.236&\cellcolor{pink!30}0.260&\cellcolor{pink!30}0.354&\cellcolor{pink!30}0.468&\cellcolor{pink!30}118&\cellcolor{pink!30}0.331&\cellcolor{pink!30}0.418  &\cellcolor{pink!30}0.507&\cellcolor{pink!30}0.622&\cellcolor{pink!30}39&\cellcolor{pink!30}0.487\\
& \textbf{SliCE}&\cellcolor{pink!30}0.185&\cellcolor{pink!30}0.250&\cellcolor{pink!30}0.331&\cellcolor{pink!30}359&\cellcolor{pink!30}0.237&\cellcolor{pink!30}0.261&\cellcolor{pink!30}0.354&\cellcolor{pink!30}0.469 &\cellcolor{pink!30}117&\cellcolor{pink!30}0.332&\cellcolor{pink!30}0.420&\cellcolor{pink!30}0.510&\cellcolor{pink!30}0.622&\cellcolor{pink!30}38&\cellcolor{pink!30}0.490\\
& \textbf{SimGCL}&\cellcolor{pink!60}0.344&\cellcolor{pink!60}0.415&\cellcolor{pink!60}0.502&\cellcolor{pink!60}\underline{162}&\cellcolor{pink!60}0.399&0.000&\cellcolor{pink!30}0.255&\cellcolor{pink!30}0.467&\cellcolor{pink!45}59&\cellcolor{pink!15}0.167&0.000&\cellcolor{pink!15}0.399&\cellcolor{pink!30}0.645&\cellcolor{pink!45}22&\cellcolor{pink!15}0.239\\
&{\textbf{RAKGE}}&\cellcolor{pink!60}\underline{0.395}&\cellcolor{pink!60}\underline{0.455}&\cellcolor{pink!60}\underline{0.529}&\cellcolor{pink!45}199&\cellcolor{pink!60}\underline{0.442}&\cellcolor{pink!60}\underline{0.502}&\cellcolor{pink!60}\underline{0.608}&\cellcolor{pink!60}\underline{0.674}&\cellcolor{pink!60}\underline{37}&\cellcolor{pink!60}\underline{0.573}&\cellcolor{pink!60}\underline{0.647}&\cellcolor{pink!60}\underline{0.733}& \cellcolor{pink!60}\underline{0.823}&\cellcolor{pink!60}\underline{12}&\cellcolor{pink!60}\underline{0.708}\\

\hline\hline
\multicolumn{2}{c|}{\textbf{NumCoKE (ours)}}&\cellcolor{pink!60}\textbf{0.439}&\cellcolor{pink!60}\textbf{0.536}&\cellcolor{pink!60}\textbf{0.640}&\cellcolor{pink!60}\textbf{119}&\cellcolor{pink!60}\textbf{0.508}&\cellcolor{pink!60}\textbf{0.636}&\cellcolor{pink!60}\textbf{0.694}&\cellcolor{pink!60}\textbf{0.758}&\cellcolor{pink!60}\textbf{30}&\cellcolor{pink!60}\textbf{0.680}&\cellcolor{pink!60}\textbf{0.745}&\cellcolor{pink!60}\textbf{0.818}&\cellcolor{pink!60}\textbf{0.888}&\cellcolor{pink!60}\textbf{7}&\cellcolor{pink!60}\textbf{0.794}\\
\hline\hline
\multicolumn{2}{c|}{lmprovement}&11.1\%&17.8\%&21.0\%&26.5\%&14.9\%&26.7\%&14.1\%&12.5\%&18.9\%&18.7\%&15.1\%&11.6\%&7.9\%&41.7\%&12.1\%\\
\hline
\end{tabular}
\caption[c]{Results for numerical reasoning. Bold scores indicate the best results, while underlined scores represent the second-best results. The \% of Improvement column shows the relative improvements of NumCoKE compared to the second-best scores.}
\label{Table2}
\end{table*}

\subsection{Theoretical Analysis}
To clarify how NumCoKE differs from other models, we compare it with popular models in the Appendix.

\noindent\textbf{Time Complexity Analysis.}
NumCoKE consists of three components, the MoEKA encoder, a process of contrastive learning, and the score function. 
In terms of MoEKA Encoder, the time complexity of processing every triple is $\mathcal{O}(\vert\mathcal{M}\vert)$. The time complexity of the numerical value embedding layer is $\mathcal{O}(\vert\mathcal{M}\vert)$ because we transform each attribute field into an embedding vector. In the joint perceptual attention layer, the query is the target relation and the number of keys and values corresponds to the number of attribute fields, so the time complexity remains $\mathcal{O}(\vert\mathcal{M}\vert)$. The gating layer of the MoEKA Encoder is not involved in the calculation of time complexity.
During the process of contrastive learning, for every given triple\textit{(h,r,t)}, the time complexity is proportional to the number of positive samples $|\mathcal{P}[h,r]|$ and the number of negative samples $\vert\mathcal{N}[h,r]\vert$. The time complexity is $\mathcal{O}(\vert\mathcal{M}\vert \cdot rel\_{}num)$, where \textit{rel\_{}num} denotes the number of relations per entity.
The TransE score function involves only addition and subtraction operations, so it does not change the time complexity.
Therefore, the overall time complexity of NumCoKE is $\mathcal{O}(\mathcal{T} \cdot (\vert\mathcal{M}\vert+ \vert\mathcal{M}\vert \cdot rel\_{}num)) \approx \mathcal{O}(\vert\mathcal{T}\vert \cdot rel\_{}num)$, where $\mathcal{T}$ denotes the number of positive and negative triples used for training.

\noindent\textbf{The Expressiveness of Our Model.} We conduct the expressiveness of numerical reasoning from the perspective of a one-hop reasoning task. Specifically, we aim to theoretically analyze the process of contrastive learning in NumCoKE and establish theoretical guarantees for the downstream performance of the learned representations. In the numerical reasoning, we denote the set of node representations as $\mathcal{V}$, and define the mean representations from one-hop neighborhoods of node $v$ as $\bm{z}  = \frac{1}{\mathcal{N}(v)}\sum\nolimits_{\bm{u} \in \mathcal{N}(v)}\bm{u}$, where $\bm{u}$ represents the projected node representations of positive samples of node $v$, $\bm{z}$ describes the one-hop neighborhood pattern of node $v$. In the world, nodes belonging to the same semantic class tend to have similar neighborhood patterns, so $\bm{z}$ can be viewed sampled from $Z|Y \sim \mathcal{N}(\textbf{z}_{Y}, I)$ where $Y$ is the latent semantic class indicating the one-hop pattern of node $v$. We demonstrate that minimizing NumCoKE’s objective in Equation (\ref{EQ20}) with an exponential moving average is equivalent to maximizing mutual information between representation $V$ and the one-hop pattern $Y$, which explains the rationality of NumCoKE in capturing one-hop patterns as follows:
\begin{equation}
    \begin{split}
        \mathcal{L}_{CL} \geq H(V \vert Y) - H(V) = -I(V;Y)
    \end{split}
    \label{LOSS-2-Theoretical}
\end{equation}
Eq. (\ref{LOSS-2-Theoretical}) indicates minimizing NumCoKE loss in Eq. (\ref{EQ20}) promotes maximizing the mutual information $I(V;Y)$ between representations and one-hop neighborhood context. In this way, our NumCoKE can exploit more latent semantics among KGs, hence facilitating numerical reasoning. All the detailed proofs are in Appendix B.

\section{Experiment}
\subsection{Experimental Settings}
\noindent\textbf{Datasets.} We select three real-world KG datasets. Specifically, US-Cities\footnote{https://simplemaps.com/data/us-cities} contains basic information about cities in the United States. 
Spotify is a dataset of songs for developers\footnote{https://www.kaggle.com/datasets/geomack}. 
Credit \cite{DBLP:journals/eswa/YehL09a} is a knowledge graph constructed from credit events in Taiwan. Additionally, all the three datasets are modified to simulate numerical reasoning tasks under real conditions. Close to 20\% of the numerical values are masked to zero (missing values). More details and the statistics of datasets can be found in Appendix C.

\noindent\textbf{Evaluation Metrics.}
We used the improved learning methods based on dropout, batch normalization, and linear transformation proposed alongside LTE \cite{DBLP:conf/www/Zhang0YW22} for evaluation. Mean Reciprocal Rank (MRR), Mean Rank (MR) and Hits@\{1, 3, 10\} are reported as metrics. Higher Hit@n and MRR values indicate a better performance, whereas lower MR values imply a better performance.

\noindent\textbf{Baseline.}
We consider six groups of 23 baseline methods: Euclidean KGE including TransE \cite{DBLP:conf/nips/BordesUGWY13}, ConvE \cite{DBLP:conf/aaai/DettmersMS018}, TuckER \cite{DBLP:conf/emnlp/BalazevicAH19}, HAKE \cite{DBLP:conf/aaai/ZhangCZW20}, 
DaBR \cite{DBLP:conf/coling/WangLBG25}, KGDM \cite{DBLP:conf/aaai/LongZLWLW24}, and RuIE \cite{DBLP:conf/coling/LiaoDH025}.
Hyperbolic KGE including MuRP \cite{DBLP:conf/nips/BalazevicAH19}, ConE \cite{DBLP:conf/nips/BaiYRL21}, AttH \cite{DBLP:conf/acl/ChamiWJSRR20}, 
GIE \cite{DBLP:conf/aaai/CaoX0CH22} and LorentzKG \cite{DBLP:conf/acl/Fan0CCDY24}; GNNs-KGE including R-GCN \cite{DBLP:conf/esws/SchlichtkrullKB18} and WGCN \cite{DBLP:conf/aaai/ShangTHBHZ19}; Attributed KGE including KBLRN \cite{DBLP:conf/uai/Garcia-DuranN18}, MT-KGNN \cite{DBLP:conf/cikm/TayTPH17} and LiteralE \cite{DBLP:conf/semweb/KristiadiKL0F19}; Self-supervised model for graph including BiGI \cite{DBLP:conf/wsdm/CaoLGLL021}, SLiCE \cite{DBLP:conf/www/WangAHCR21}, SimGCL \cite{DBLP:conf/sigir/YuY00CN22}, and 
RAKGE \cite{DBLP:conf/kdd/KimKKPJP23}; Deterministic value representation methods including NEKG \cite{DBLP:conf/emnlp/DuanYT21} and 
NRN \cite{DBLP:conf/kdd/BaiLLYYS23}. Among all the baselines, for those that do not take care of external numerical attributes, such as TransE, we follow the original approach and concat the embeddings of entity and attributes. The results of self-supervised KGE are from RAKGE. More details can refer to the Appendix C.


\subsection{Experimental Results}
The main results are detailed in Table\ref{Table2}. We conduct 5 rounds of experiments and take the average as the final result, which reveals that NumCoKE makes significant progress in all the metrics and achieves new SOTA results. All the evaluated models are implemented on a server with one GPU (\textit{NVIDIA A800, 80GB}).
Notably, MRR on Credit increased from 0.493 for TransE and 0.498 for DaBR to 0.794, which highlights NumCoKE's robust capability in capturing the asymmetric nature of relations and the ordinal information. At the same time, NumCoKE's MRR significantly outperforms to hierarchy-aware models such as TuckER and hyperbolic KGE methods, affirming its superior utilization of the hierarchical structures inherent in numerical reasoning relations. 
RuIE performs weakly, which we infer is due to excessive focus on logical rules rather than comparative relationships between numerical semantics.
Although utilizing the extra attributes of entities, attribute-based models like KBLRN and MT-KGNN experience performance declines due to the relevance of numerical attributes, whereas NumCoKE excels under these conditions. 
Deterministic approaches such as NEKG and NRN are unable to recognize semantic differences between numerical fields and only consider the distribution of attribute values in each field (e.g., \textit{51kg} and \textit{51 years old}), whereas NumCoKE sets up different representation for each field and utilizes the \textit{knowledge perceptual attention mechanism} that allows it to focus on attributes that are more relevant to the relation. 
In addition, compared to LiteralE and RAKGE, NumCoKE excels by fully exploiting the potential of the relational contexts and the interactions among entities, relations and numerical attributes.  More details can refer to the study for ordinal knowledge contrastive learning in Appendix D.

\subsection{Experimental Analysis}
\noindent\textbf{Ablation Study.} 
To assess the contribution of each component in \textbf{NumCoKE}, we conduct ablation studies by removing key modules individually.
(1) Knowledge-aware attention (KA): 
We disable the KA module by setting $\delta_e = \delta_r = 0$ in Eq.~(\ref{EQ Attention}), effectively removing the influence of relation- and entity-specific signals on attribute modeling. As shown in Table~\ref{Table3}, this leads to a significant performance drop (↓8.1\% on Credit), confirming the importance of incorporating both entity and relation semantics for numerical reasoning.
(2) Mixture-of-experts (MoE):  
We remove the MoE mechanism and instead use static entity representations across all relation contexts. This results in a 5.0\% accuracy drop, indicating that dynamically adapting entity embeddings based on relational context is crucial for capturing semantic nuances.
(3) Ordinal sampling (OS):  
We further ablate the top-$k$ ordinal sampling strategy used in Eq.~(\ref{sample positive})–(\ref{sample negative}). Without OS, the model relies solely on randomly sampled contrastive pairs, resulting in degraded performance (↓6.4\% on Credit). This highlights the effectiveness of OS in improving the model’s ability to distinguish subtle ordinal differences among close numerical values. We also conduct a single-module-addition study, the results can be seen in the Appendix.

\begin{table}[htbp]
\centering
\footnotesize
\setlength{\tabcolsep}{5pt}
\begin{tabular}{c|c|c|ccccc}
\hline\hline
\bf{MoE}&\bf{KA}&\bf{OS}  & \bf{H@1} & \bf{H@3} & \bf{H@10} & \bf{MR} & \bf{MRR} \\
\hline
$\times$&$\surd$&$\surd$&0.712&0.764&0.851&9& 0.765\\
$\surd$&$\times$&$\surd$&0.688&0.763&0.846&10&0.740\\
$\surd$&$\surd$&$\times$&0.698&0.776&0.843&10&0.750\\
\hline
\rowcolor[gray]{0.9}
$\surd$&$\surd$&$\surd$&\bf{0.745}&\bf{0.818}&\bf{0.888}&\bf{7}&\bf{0.794}\\
\hline\hline    
\end{tabular}
\caption[c]{Ablation study on Credit dataset.}
\label{Table3}
\end{table}

\begin{figure}[htbp]
    \centering
    \includegraphics[scale=0.165]{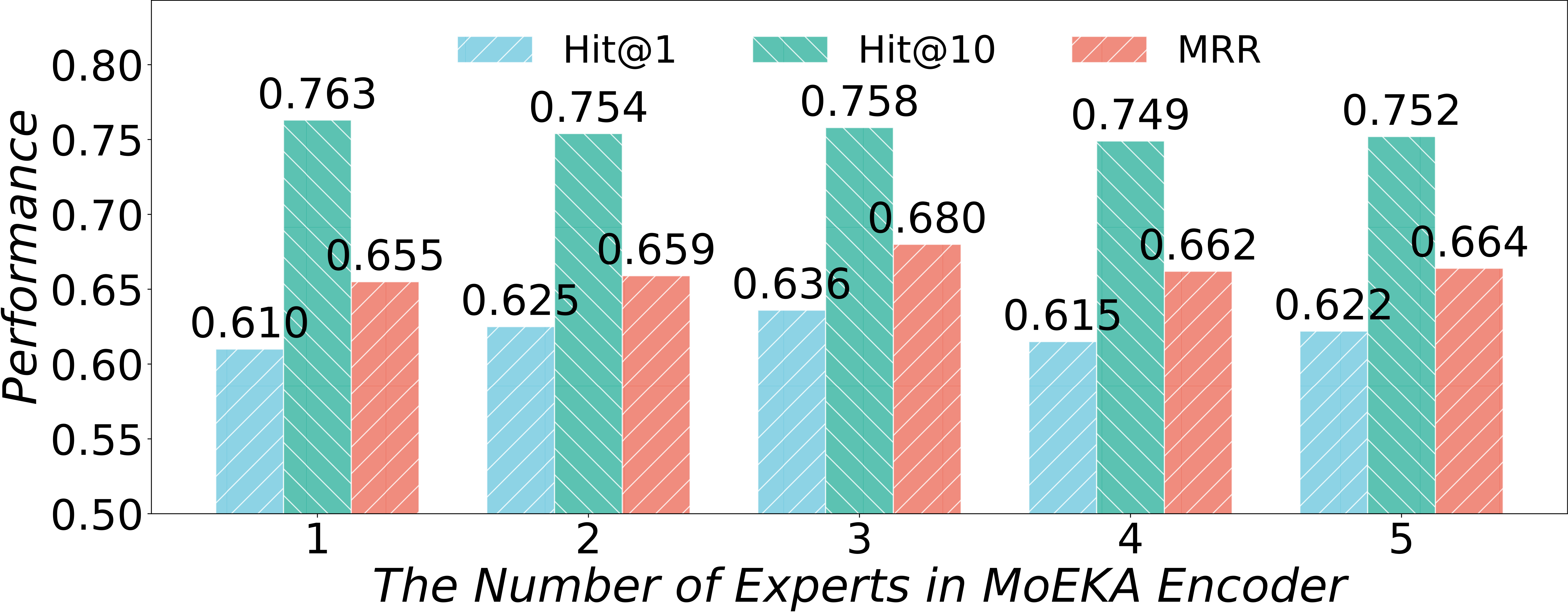}
    \caption{Research of Hyperparameter \textit{K} on Spotify dataset.}
    \label{MoE-Experts}
\end{figure}

\noindent\textbf{Analysis of MoEKA Encoder.} 
We investigate the effect of the number of experts \textit{K} in the MoE module, as depicted in Figure \ref{MoE-Experts}. It can be observed that the impact of the number of experts \textit{K} on the final results generally follows a pattern of initial increase followed by a decrease, mainly affecting fine-grained metrics such as Hit@1 and MRR. Having either too many or too few experts is detrimental to the model’s learning performance, 
this is because the same relational category corresponds to similar contexts, for example, \textit{ranking\_{}comp} and \textit{disabled\_{}comp} are both comparative relations, and too many experts can cause semantic confusion.  

\begin{figure}[htbp]
    \centering
    \includegraphics[scale=0.155]{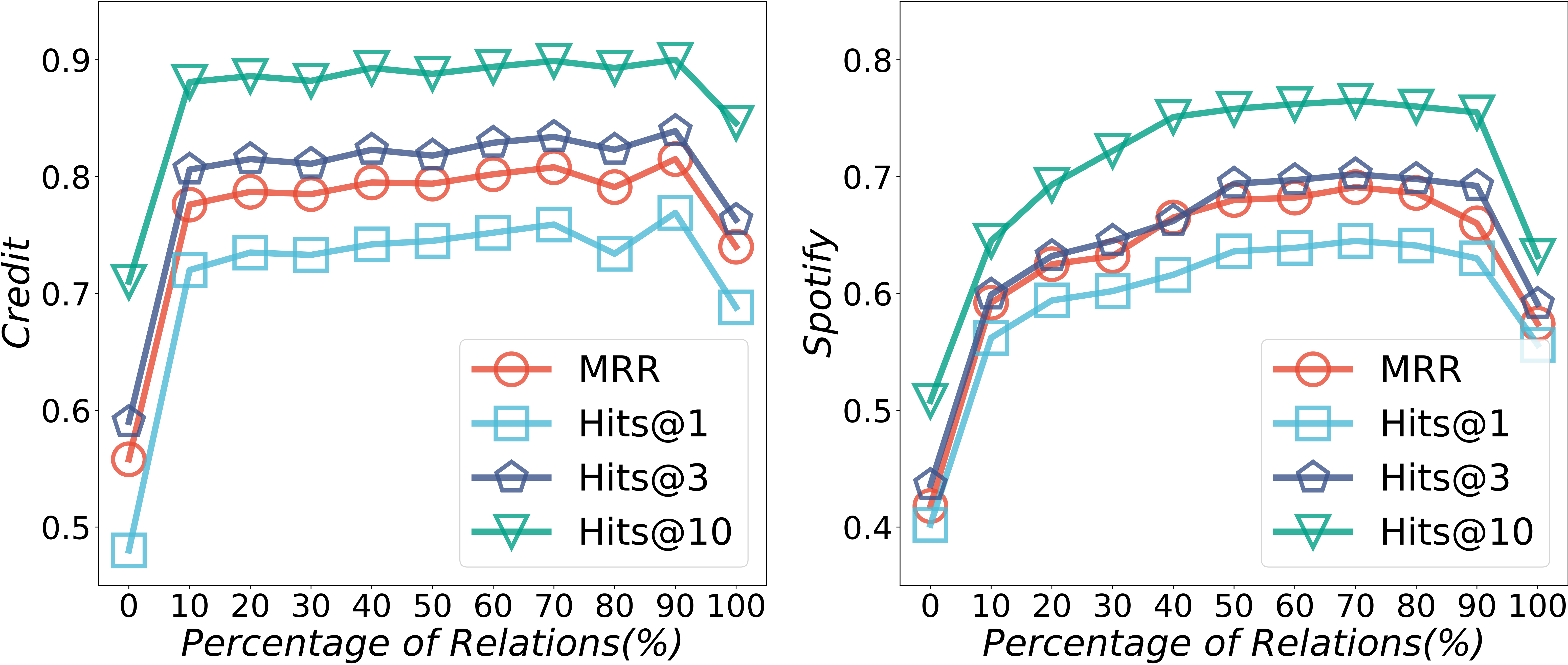}
    \caption{Proportional test on Credit and Spotify. A relation percentage of 100\% means that we only consider relations, while 0\% means that we only consider entities.}
    \label{Percentage}
\end{figure}

\noindent\textbf{Effect of Relation-Entity Perception Weights.}
To assess the relative contributions of entities and relations in numerical attribute modeling, we vary their attention weights in the joint perception mechanism (Eq.~\ref{EQ Attention}). Specifically, we adjust the proportion assigned to relations from 0\% to 100\% and evaluate the performance of the model on the \textsc{Credit} dataset.
As shown in Figure~\ref{Percentage}, \textbf{NumCoKE achieves optimal performance when relations contribute 40\%–90\% of the total attention weight}, with peak results at 90\%. Compared to the entity-only variant (0\% relation weight), this configuration improves MRR by \textbf{46.1\%} (from 0.186 to 0.272), demonstrating the significant value of relation semantics in attribute perception.
However, performance deteriorates when relations are used exclusively (i.e., 100\%), mirroring the trend observed in RAKGE~\cite{DBLP:conf/kdd/KimKKPJP23}, which neglects entity semantics. These results \textbf{highlight the necessity of jointly modeling entity and relation semantics} to effectively capture numerical attributes.



\begin{figure}[htbp]
    \centering
    \includegraphics[scale=0.090]{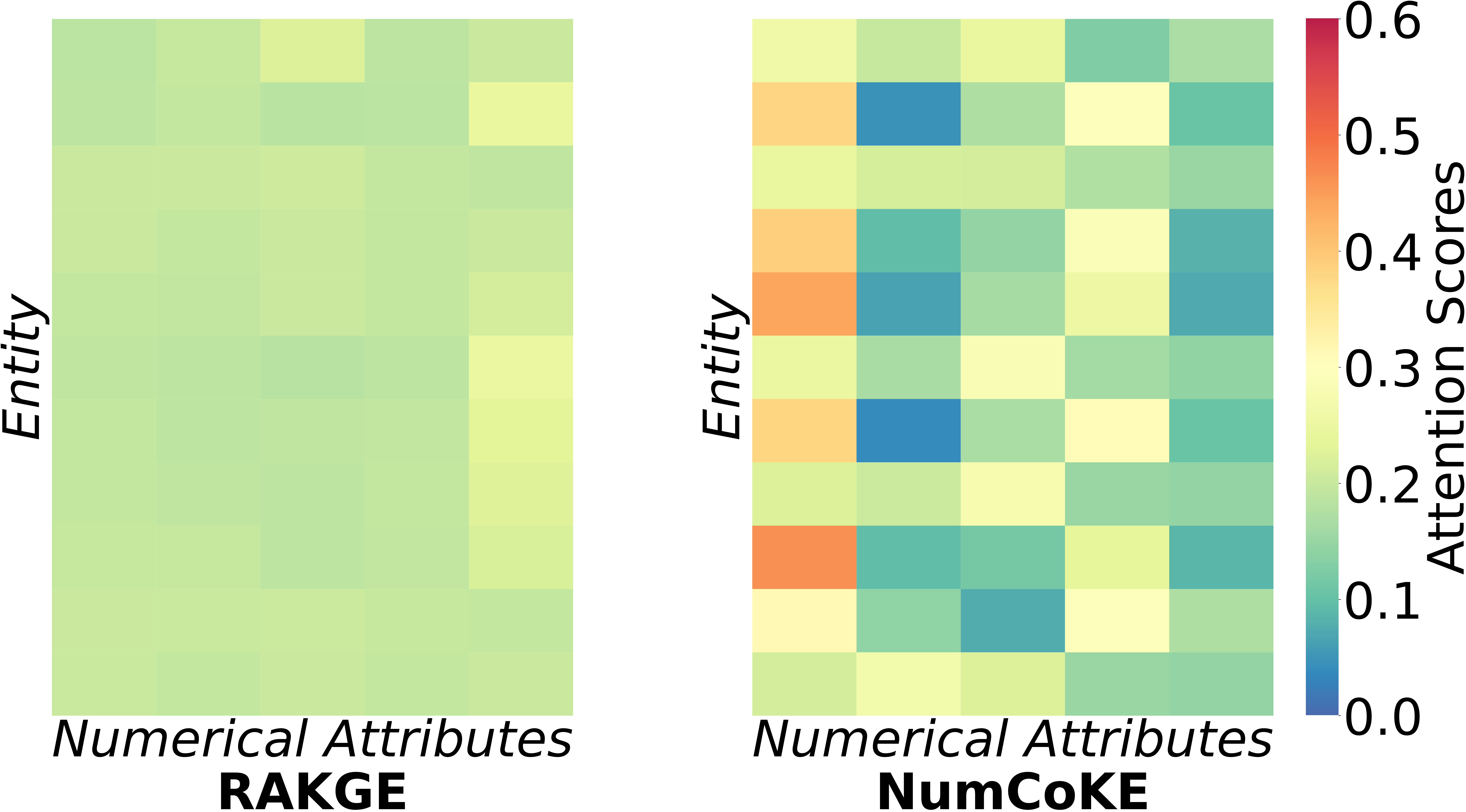}
    \caption{Visualization of relevance scores of each numeric attribute of RAKGE and NumCoKE on Credit dataset.}\label{Heatmap}
\end{figure}

\noindent\textbf{Case Study.} Furthermore, for easier visualization, we select part of the samples from Credit to show the attention scores between entities and attributes using heatmaps. As shown in Figure \ref{Heatmap}, the color blocks in each row of the first heat map are similar, indicating that RAKGE pays almost equal attention to each attribute in most samples, while the color blocks from NumCoKE are clearly differentiated, demonstrating that NumCoKE can better distinguish the importance of attributes in different relation contexts. More case studies can refer to Appendix E for more details.

\subsection{Supplementary Experiments}
We performed other experiments, including hyperparameter analysis, evaluation on FB15k-237, etc. (see Appendix E).

\section{Conclusion}
In this paper, we introduce a novel KGE model called NumCoKE. Specifically, we incorporates a Mixture-of-Experts-Knowledge-Aware Encoder for elaborate semantic modeling, capturing the connections among entities, relations, and attributes within a unified framework. To distinguish similar semantics, we introduce a novel ordinal knowledge contrastive learning strategy that that generates high-quality ordinal samples from original data, capturing fine-grained semantic nuances over close numerical values. Extensive experiments on three standard benchmarks demonstrate NumCoKE's effectiveness for numerical reasoning on KGs.

\section{Acknowledgments}
This work was supported by the National Key R\&D Program of China under Grant 2022YFB3903904, the Key R\&D Program of Jiangxi Province under Grant 20232BBGW001.


\bigskip
\bibliography{sample-base}

\end{document}